\documentclass[runningheads]{llncs}
\usepackage[T1]{fontenc}
\usepackage{graphicx}
\usepackage{amsmath}
\usepackage{booktabs}

\usepackage{multirow}
\usepackage{tabularx}
\usepackage{tikz}
\usetikzlibrary{positioning,fit,backgrounds,calc,shadows,arrows.meta}

\begin{document}
\title{Evaluating Learner Representations for Differentiation Prior to Instructional Outcomes}
\titlerunning{Evaluating Learner Representations for Differentiation}
%
\author{Junsoo Park\orcidID{0009-0003-4450-0103} 
\and
Youssef Medhat \and
Htet Phyo Wai \and
Ploy Thajchayapong \and
Ashok K. Goel\orcidID{0000-0003-4043-0614}}
\authorrunning{J. Park et al.}
%
\institute{Georgia Institute of Technology, Atlanta, GA, USA\\
\email{jpark3232@gatech.edu, ymedhat3@gatech.edu, hwai6@gatech.edu, ploy@gatech.edu, ashok.goel@cc.gatech.edu}}
\maketitle              
\begin{abstract}
Learner representations play a central role in educational AI systems, yet it is often unclear whether they preserve meaningful differences between students when instructional outcomes are unavailable or highly context-dependent. This work examines how to evaluate learner representations based on whether they retain separation between learners under a shared comparison rule. We introduce \emph{distinctiveness}, a representation-level measure that evaluates how each learner differs from others in the cohort using pairwise distances, without requiring clustering, labels, or task-specific evaluation. Using student-authored questions collected through a conversational AI agent in an online learning environment, we compare representations based on individual questions with representations that aggregate patterns across a student’s interactions over time. Results show that learner-level representations yield higher separation, stronger clustering structure, and more reliable pairwise discrimination than interaction-level representations. These findings demonstrate that learner representations can be evaluated independently of instructional outcomes and provide a practical pre-deployment criterion using distinctiveness as a diagnostic metric for assessing whether a representation supports differentiated modeling or personalization.
\end{abstract}

\section{Introduction}
In large, asynchronous courses, evidence of student understanding is distributed across brief and fragmented interactions, limiting instructor awareness of which students require different forms of support \cite{zou2025digital,park2025human}. Educational AI systems increasingly rely on learner representations to summarize such interaction data, yet these representations remain approximations rather than ground-truth models of learning \cite{bull2016smili}. As a result, personalization is difficult to evaluate directly, particularly when outcome-based measures are unavailable during early system design \cite{hattie2007power}. This work addresses a foundational question: \emph{how can we evaluate whether learner representations preserve meaningful differences among students before instructional outcomes are observed?} We frame differentiation among learners as a necessary but not sufficient condition for personalization—--if learners cannot be distinguished within a representation, there is limited basis for reasoning about how instructional support should differ. This perspective aligns with foundational work in mastery learning, differentiated instruction, and adaptive educational systems, which treat the identification of learner differences as a prerequisite for personalization \cite{bloom1968learning,tomlinson1999differentiated,brusilovsky2001adaptive}. To operationalize this idea, we introduce \emph{distinctiveness}, a structural property that captures the extent to which learners remain differentiated under a common similarity measure. Using student-authored questions from an online learning environment, we compare interaction-level representations derived from individual questions with learner-level representations constructed from aggregated interaction histories. Through this comparison, we show how representational choices affect the preservation of learner differences. This work proposes a representation-centered approach for evaluating learner differentiation prior to instructional use.

\section{Research Question}
\vspace{-0.5em}
In online education, within adult learning contexts, conversational AI agents are used to motivate students to ask questions \cite{goel2018jill,basu2025bidirectional}. Prior work on personalized learning emphasizes that effective personalization depends on identifying meaningful differences among learners and using those differences to inform instructional decisions \cite{bernacki2021systematic}. However, evaluating such representations remains challenging due to the absence of ground-truth learner models and outcome-based measures during system design. Accordingly, this work investigates the following research question:

\textbf{RQ1:} How can learner representations be evaluated for their capacity to preserve meaningful differentiation among students?

We introduce \emph{distinctiveness}, a structural property capturing the extent to which learners remain distinguishable under consistent similarity criteria. The formulation is inspired by sensitivity in differential privacy \cite{mcsherry2009differentially}, which bounds how much outputs change when an individual’s data is modified. Here, we consider the complementary perspective: whether representations preserve differences between individuals. We further draw on notions of local similarity from $k$-anonymity \cite{samarati1998protecting,sweeney2002k} to characterize neighborhood density, where lower density reflects greater differentiation rather than a privacy guarantee. We evaluate the following hypothesis:

\textbf{H1:}
Representational evaluations that operate at the learner level, rather than at the level of individual interactions, will reveal greater distinctiveness and clearer separation among students under a common similarity measure.

\section{Background and Related Work}
\subsection{Learner Representations in Personalized Learning}
Personalized instruction has longstanding foundations in mastery learning and differentiated instruction, which emphasize responding to meaningful differences among learners \cite{bloom1968learning,tomlinson1999differentiated}. In educational AI systems, learner differences are captured through representations that summarize patterns of engagement, progress, or need. Prior work in adaptive and personalized learning has relied on learner representations to support instructional adaptation \cite{brusilovsky2001adaptive}, yet offers limited guidance on how to evaluate whether such representations preserve meaningful differentiation among learners, independent of instructional outcomes.

\subsection{Challenges in Evaluating Learner Representations}
Evaluating personalization remains challenging in educational contexts. Instructional responses are inherently context-dependent and do not admit a single correct form of feedback \cite{hattie2007power}, while learner representations provide interpretable approximations rather than ground-truth models of learning \cite{bull2016smili}. As a result, outcome-based evaluation offers limited guidance during early system design. To address this, we draw on concepts from privacy research as analytical lenses. In differential privacy, sensitivity characterizes how system outputs vary in response to changes in individual inputs \cite{mcsherry2009differentially}. Analogously, we examine whether learner representations remain responsive to individual differences, using structural properties of the representation itself rather than instructional outcomes.

\section{Study Design}
\vspace{-0.5em}
\begin{figure}[t]
    \centering
    \includegraphics[width=0.9\linewidth]{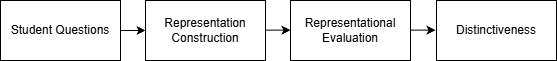}
    \caption{Conceptual framing of the evaluation. Student-authored questions are transformed into alternative representational forms, which are analyzed for the extent to which learners remain differentiated. Distinctiveness serves as an outcome-independent proxy for representational differentiation.}
    \label{fig:pipeline}
\end{figure}

As illustrated in Figure~\ref{fig:pipeline}, our evaluation compares learner representations constructed from the same interaction data. Student-authored questions serve as the common input. From this shared input, we construct two representations that differ in how learner information is aggregated, and evaluate how well each preserves differentiation.

\subsection{Units of Representation}
We represent each learner with a fixed-length vector $\mathbf{s}_i$ in one of two ways: as the mean of embeddings of that learner’s questions (interaction-level), or as aggregated features from the learner’s questions and interaction history (learner-level). In both cases, $\mathbf{s}_i$ is a fixed-length summary of the learner’s interaction history. We compare two representational formats:

\begin{itemize}
    \item \textbf{Interaction-level (question embedding) representations.} Each question is encoded as a 384-D vector using the Sentence Transformers library\footnote{\url{https://www.sbert.net/}} with the \texttt{all-MiniLM-L6-v2} checkpoint\footnote{\url{https://huggingface.co/sentence-transformers/all-MiniLM-L6-v2}}. Learner representations are computed as the mean embedding of a learner’s questions. For pairwise verification, individual question embeddings are used directly.

    \item \textbf{Learner-level (risk signature) representations.} Each learner is represented by a 45-D vector constructed from aggregated interaction signals, including summary statistics of instructional-need scores, recommendation text embeddings (using the same encoder), temporal features, and aggregated question embedding features. All features are min--max normalized per dimension across learners prior to distance computation.
\end{itemize}

\begin{figure}[t]
    \centering
    \includegraphics[width=0.65\linewidth]{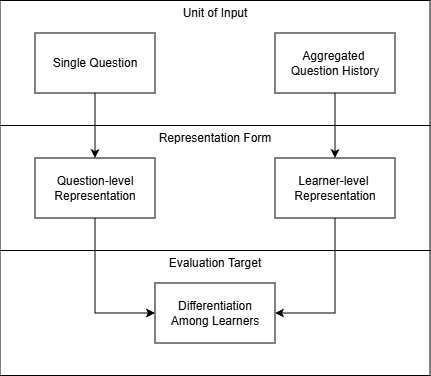}
    \caption{Conceptual framing of the comparison. Representations derived from individual questions and from aggregated question histories are evaluated with respect to the same objective: the degree to which learners remain differentiated.}
    \label{fig:framing}
\end{figure}

Both representations are derived from the same underlying questions. Distinctiveness and learner uniqueness are evaluated using normalized L2 distance (Eq.~\ref{eq:distinctiveness}), while complementary indicators use their respective distance measures as defined in Section~\ref{sec:indicators}. For pairwise verification, interaction-level representations use per-question embeddings, while learner-level representations use per-interaction instantiations of learner signatures to construct same- and cross-learner pairs.

\subsection{Participants and Data}
The analysis uses anonymized interaction logs from an AI-based virtual teaching assistant deployed in a graduate computer science course at a U.S.\ R1 research university \cite{goel2018jill,goel2025a4l,thajchayapong2025evolution}. Students ask questions during coursework, including articulations of uncertainty and reasoning. The dataset contains 8{,}838 student-authored questions from 200 learners collected over a semester. Questions are open-ended and heterogeneous (e.g., conceptual, procedural, debugging, reflection). We do not score questions for correctness or quality, as the evaluation focuses on representation structure rather than labeled outcomes. The analysis uses only interaction-derived representations, excluding demographic, performance, and outcome data. Reported results include $n{=}39$ learners with complete interaction histories.

\subsection{Analysis of Representational Differentiation}
We evaluate whether learner-level representations produce greater separation between learners than interaction-level representations derived from individual questions. This comparison is conducted under controlled conditions: both representations are constructed from the same underlying questions and evaluated using consistent procedures for each metric. The analysis focuses on representation structure, independent of instructional outcomes.

\subsubsection{Distinctiveness}
\label{sec:distinctiveness}
We define \emph{distinctiveness} as the mean distance between a learner’s representation and those of all other learners, using a normalized L2 metric that divides distances by $\sqrt{d}$ to enable comparison across representations of different dimensionality:

\begin{equation}
\label{eq:distinctiveness}
D_{\text{norm}}(i)=\frac{1}{N-1}\sum_{j\neq i}\frac{\|\mathbf{s}_i-\mathbf{s}_j\|_2}{\sqrt{d}}.
\end{equation}

This computes, for each learner, the mean distance to all others, normalized by dimensionality. Here $\mathbf{s}_i$ is the representation of learner $i$, $d$ is the dimensionality, and $N$ is the number of learners. Higher values indicate greater separation from other learners. We report the mean and standard deviation of \(D_{\text{norm}}(i)\) across learners. Distinctiveness serves as our primary measure of learner-level differentiation: it summarizes separation without requiring clustering, labels, or instructional outcome data.

\subsubsection{Complementary Separation Indicators}
\label{sec:indicators}
To contextualize distinctiveness, we report three complementary indicators that capture distinct aspects of learner separation:

\begin{enumerate}
    \item \textbf{Overall differentiation.}
    We report the silhouette coefficient, which compares how close each learner is to others in the same $k$-means cluster versus to learners in other clusters. Cluster labels are obtained using scikit-learn $k$-means, with the number of clusters set as a function of sample size and bounded to avoid extreme values. Silhouette is then computed on the same representations using Euclidean distance\footnote{\url{https://scikit-learn.org/stable/modules/generated/sklearn.metrics.silhouette\_score.html}}.

    \item \textbf{Pairwise verification.}
    We evaluate whether representations from the same learner are more similar than those from different learners using cosine similarity. The procedure is as follows:
    \begin{itemize}
        \item \textit{Pair construction:} Positive pairs consist of two representations from the same learner, and negative pairs consist of representations from different learners, with matched cardinality.
        \item \textit{Scoring:} For each pair, we compute cosine similarity.
        \item \textit{Evaluation:} ROC--AUC is computed from the similarity scores using the scikit-learn implementation\footnote{\url{https://scikit-learn.org/stable/modules/generated/sklearn.metrics.roc\_auc\_score.html}}.
    \end{itemize}
    ROC--AUC summarizes how well similarity scores separate positive and negative pairs. It can be interpreted as the probability that a randomly selected same-learner pair has higher similarity than a randomly selected cross-learner pair.

    \item \textbf{Learner uniqueness.}
    We evaluate how many other learners fall within a given threshold $\tau$ under normalized L2 distance. We then increase $\tau$ and identify the smallest value at which even the most isolated learner has at least one neighbor within the threshold. This threshold is used only to compare how quickly learners become non-unique under different representations.
\end{enumerate}

\subsection{Results}
\label{sec:results}
\vspace{-0.5em}
\begin{table}[t]
\centering
\caption{Representational differentiation for interaction-level and learner-level representations (\(n=39\) students).}
\label{tab:main_results}
\footnotesize
\setlength{\tabcolsep}{3pt}
\begin{tabular}{lcccc}
\toprule
\textbf{Representation}
& \(\mathbf{D}\) (mean\(\pm\)SD)
& \(\mathbf{S}\)
& \(\mathbf{A}\)
& \(\boldsymbol{\tau_{k>1}}\) \\
\midrule
Question-level
& \(1.072 \pm 0.063\)
& 0.028
& 0.626
& 0.052 \\

Learner-level
& \(1.435 \pm 0.093\)
& 0.507
& 0.878
& 0.3409 \\
\bottomrule
\end{tabular}

\begin{flushleft}
\scriptsize
\textit{Note.} \(D\) is per-student distinctiveness (Eq.~\ref{eq:distinctiveness}). \(S\): silhouette (partition-dependent). \(A\): pairwise verification ROC--AUC. \(\tau_{k>1}\): smallest normalized distance at which all learners have at least one neighbor (i.e., no learner remains unique). Larger implies more robust differentiation. Threshold semantics remain interpretively open.
\end{flushleft}
\end{table}

Learner-level representations exhibited substantially stronger differentiation than interaction-level question embeddings (Table~\ref{tab:main_results}). Learner-level distinctiveness was higher ($1.435 \pm 0.093$) than question-level distinctiveness ($1.072 \pm 0.063$), reflecting a 34\% increase in average separation ($n=39$). This is supported by complementary measures: learner-level representations formed coherent groupings (silhouette $=0.507$ vs.\ $0.028$) and more reliably placed the same learner closer to themselves than to others under pairwise verification (ROC--AUC $=0.878$ vs.\ $0.626$). Learner-level representations also remained distinguishable at larger distance thresholds, with loss of uniqueness occurring at $\tau_{k>1}=0.3409$ compared to $0.052$ for question-level embeddings.

\section{Discussion}
\subsection{What Representations Reflect}
Learner-level differentiation depends on whether students remain distinguishable under a common similarity measure. When learners appear similar, differences become difficult to interpret. Our results show that interaction-level question embeddings lose this separation quickly: as similarity criteria are broadened, students who ask overlapping questions become harder to distinguish. In contrast, learner-level representations preserve separation under broader comparisons, even when individual questions overlap. Interaction-level representations emphasize semantic overlap in isolated question text, whereas learner-level signatures aggregate signals across learner activity (e.g., instructional needs, temporal patterns, and aggregated embedding features). As a result, comparisons shift from individual questions to broader patterns, revealing distinctions not apparent from isolated interactions. These patterns are consistent across all separation indicators reported in Section~\ref{sec:results}, including distinctiveness, silhouette score, and pairwise ROC--AUC.

\subsection{What Distinctiveness Measures}
\emph{Distinctiveness} (Eq.~\ref{eq:distinctiveness}) measures the average distance between each learner and all others. It evaluates whether learners remain distinguishable under a shared distance rule and should be interpreted as a property of the representation rather than a model of the learner. Separation is treated as a necessary but not sufficient condition: if learners are not distinguishable, there is no basis for differentiated reasoning, though large separation may also reflect noise. Distinctiveness does not distinguish between meaningful variation and variation introduced by feature construction, and interpreting differences requires external validation. Compared to complementary indicators, distinctiveness evaluates separation without imposing additional structure. Silhouette depends on clustering and a choice of $k$, while ROC--AUC evaluates a pairwise classification task. Distinctiveness instead compares each learner to the full cohort, incorporating all pairwise relationships without clustering or sampling. Unlike dispersion measures such as variance, which summarize spread relative to a central point, distinctiveness measures separation directly between learners. This aligns with instructional settings in which learners are interpreted relative to others in the classroom rather than to an abstract average \cite{tomlinson1999differentiated}. Distinctiveness summarizes average separation, while uniqueness reflects the most isolated learner. In our results, learner-level representations improved both, indicating stronger separation both globally and locally. Future work should examine how feature composition and dimensionality contribute to these differences.

\subsection{Implications for Personalization and Learner Modeling}
Distinctiveness indicates whether a representation provides the information necessary for differentiated reasoning about learners. If learners are not separated, there is little basis to treat them differently, consistent with prior work emphasizing that personalization depends on identifying meaningful learner differences \cite{bernacki2021systematic,bloom1968learning,tomlinson1999differentiated}. It can identify representations that collapse learners into similar profiles and those that retain variation across learners. When external measures are available (e.g., assessments or surveys), such representations are more likely to contain variation that relates to those measures, though this must be verified empirically. In practice, these representations can serve as inputs to standard modeling approaches (e.g., clustering, classification, regression). For example, a practitioner could compare candidate representations and select those that preserve greater separation before training models for personalized learning \cite{bernacki2021systematic,brusilovsky2001adaptive}. This does not establish instructional meaning, but ensures that the representation retains variation that could support personalization.

\section{Conclusion}
This work examined how different ways of representing learners affect which differences between students remain visible prior to instructional evaluation. Using student-authored questions from an online classroom, we compared representations based on individual questions with representations that aggregate patterns across a learner’s interactions over time. We introduce \emph{distinctiveness} as a task-independent measure of whether learner representations preserve between-learner separation prior to outcome-based validation. Unlike clustering- or task-based approaches, distinctiveness evaluates separation directly across the full cohort under a shared comparison rule, without imposing group structure or reducing learners to a central reference point. In this sense, it captures how each learner stands out relative to others in the cohort. We further show empirically that learner-level aggregation improves this property relative to interaction-level representations. Across all analyses, learner-level representations exhibited higher distinctiveness, silhouette scores, and pairwise ROC--AUC than interaction-level representations, supporting H1. Distinctiveness does not require instructional labels or performance data and can be applied to representations constructed from interaction traces. Conceptually, it is grounded in a simple geometric principle: when no grouping or reference point is assumed, differences between learners must be defined relative to one another. Distinctiveness operationalizes this by comparing each learner to the full cohort under a shared distance measure. In practice, this provides a pre-deployment criterion: without separation between learners, there is limited basis for differentiated modeling or personalization.

\subsubsection{Acknowledgements} This research has been supported by grants from the US National Science (\#2247790 and \#2112532) to the National AI Institute for Adult Learning and Online Education (aialoe.org).

%
%
%
\bibliographystyle{splncs04}
\bibliography{mybibliography}

\end{document}